\documentclass[letterpaper]{article} 
\usepackage{aaai25}  
\usepackage{times}  
\usepackage{helvet}  
\usepackage{courier}  
\usepackage[hyphens]{url}  
\usepackage{graphicx} 
\usepackage{natbib}  
\usepackage{caption} 
\usepackage{algorithm}
\usepackage{algorithmic}
\usepackage{amssymb}
\usepackage{amsmath}
\usepackage{amsthm}
\usepackage{booktabs}
\usepackage{multirow}
\usepackage{graphicx} 
\usepackage{subfigure}
\usepackage{times}

\usepackage{xcolor}

\usepackage{newfloat}
\usepackage{listings}
\DeclareCaptionStyle{ruled}{labelfont=normalfont,labelsep=colon,strut=off} 
\floatstyle{ruled}
\newfloat{listing}{tb}{lst}{}
\floatname{listing}{Listing}
\title{SCFCRC: Simultaneously Counteract Feature Camouflage and Relation
Camouflage for Fraud Detection}

\author{
    Xiaocheng Zhang\textsuperscript{\rm 1,3}\equalcontrib,
    Zhuangzhuang Ye\textsuperscript{\rm 1}\equalcontrib,
    GuoPing Zhao\textsuperscript{\rm 2},
    Jianing Wang\textsuperscript{\rm 3},
    Xiaohong Su\textsuperscript{\rm 1}
}
\affiliations{
    \textsuperscript{\rm 1}Faculty of Computing, Harbin Institute of Technology\\
    \textsuperscript{\rm 2}Zhengzhou Esunny Information Technology Co., Ltd.,
    \textsuperscript{\rm 3}MeiTuan\\
    22S136029@stu.hit.edu.cn, 22S136019@stu.hit.edu.cn, sxh@hit.edu.cn
}

\usepackage{bibentry}

\begin{document}

\maketitle

\begin{abstract}
In fraud detection, fraudsters often interact with many benign users, camouflaging their features or relations to hide themselves. Most existing work concentrates solely on either feature camouflage or relation camouflage, or decoupling feature learning and relation learning to avoid the two camouflage from affecting each other. However, this inadvertently neglects the valuable information derived from features or relations, which could mutually enhance their adversarial camouflage strategies.
    In response to this gap, we propose SCFCRC, a Transformer-based fraud detector that \textbf{S}imultaneously \textbf{C}ounteract \textbf{F}eature \textbf{C}amouflage and \textbf{R}elation \textbf{C}amouflage. SCFCRC consists of two components: Feature Camouflage Filter and Relation Camouflage Refiner. The feature camouflage filter utilizes pseudo labels generated through label propagation to train the filter and uses contrastive learning that combines instance-wise and prototype-wise to improve the quality of features. The relation camouflage refiner uses Mixture-of-Experts(MoE) network to disassemble the multi-relations graph into multiple substructures and divide and conquer them to mitigate the degradation of detection performance caused by relation camouflage. Furthermore, we introduce a regularization method for MoE to enhance the robustness of the model. Extensive experiments on two fraud detection benchmark datasets demonstrate that our method outperforms state-of-the-art baselines.
\end{abstract}

%

\section{Introduction}
\begin{figure}[t]
\centering
\includegraphics[width=0.45\textwidth]{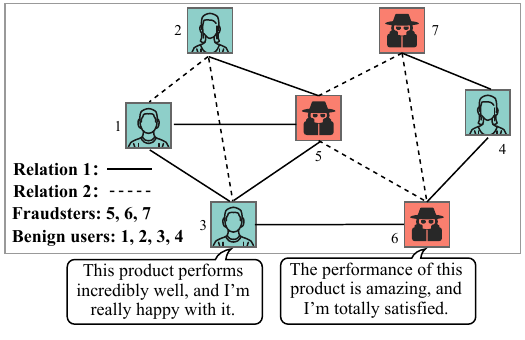}
\caption{A toy example illustrating camouflaged fraud: (1) Feature Camouflage: Fraudsters mimic benign users' behaviors, like posting reviews. For instance, fraudster "6" fakes a positive review for a product, adopting the style of normal user "3".
(2) Relation Camouflage: Under relation 1, fraudsters establishes connections with normal users to avoid suspicion. But they primarily connect with other fraudsters under relation 2, sparingly engaging with normal users.}
\label{fig:example}
\end{figure}

Along with the increasing popularity of e-commerce platforms and social media, fraud has increased dramatically.
For example, the proliferation of fake reviews on e-commerce platforms may impact both merchants and consumers \cite{li2019spam} and the presence of fraudulent accounts in financial transactions has caused huge property losses for clients\cite{liang2021credit,ren2021ensemfdet}.
In recent years, graph-based fraud detection methods have been widely applied in many practical applications.
Researchers usually model entities as nodes and interactions between entities as edges to solve the fraud problem from a graph perspective.
Due to the superior representation ability of graph neural networks(GNNs), GNN-based fraud detection methods have attracted extensive attention from industry and academia~\cite{liu2020alleviating,zhang2021fraudre,xiang2023semi}.
These methods are based on the assumption that entities with the same goal tend to have ”homophily” neighborhoods, which means the center nodes rely on information propagated from neighboring nodes in the same class.

However, to avoid being detected by fraud detection systems, fraudsters will try their best to camouflage themselves.
Common methods include mimicking the behavioral patterns of benign users and establishing connections with benign users\cite{hooi2016fraudar,ge2018securing,liu2020alleviating}.
The first type of camouflage behavior is called feature camouflage and the second type is called relation camouflage.
Figure~\ref{fig:example} provides a toy example illustrating the two types of camouflage behavior described above.
Fraudsters' camouflage behavior breaks the 'homophily' assumption, where fraudsters may interact with many benign users and have similar characteristics to benign users.
Camouflage causes features and connections to be entangled during the propagation process of GNN, making feature learning and structure learning affect each other, which may exaggerate false information\cite{meng2023decoupling}.
Many existing approaches attempt to alleviate the camouflage phenomenon, but most of them only address one type of camouflage and ignore the other. Some methods \cite{meng2023decoupling} address both types of camouflage separately and then simply integrate them while neglecting the deeper relationship.

To address these issues, we propose SCFCRC, a Transformer-based Fraud Detector that \textbf{S}imultaneously \textbf{C}ounteract \textbf{F}eature \textbf{C}amouflage and \textbf{R}elation \textbf{C}amouflage.
SCFCRC contains two components: feature camouflage filter and relation camouflage refiner.
The feature camouflage filter is to solve the feature camouflage.
We first obtain pseudo-labels for each node using the label propagation method, in which only the graph structure information is used and ignoring the original features of the nodes, thus avoiding the negative impact of feature camouflage.
The pseudo labels are then used to train the feature camouflage filter.
During this period, we use instance-wise and prototype-wise contrastive learning to improve feature quality.
Subsequently, the filtered features are sent to the relation camouflage refiner along with the original features.
Inspired by the group aggregation strategy\cite{wang2023label}, we perform group aggregation on the filtered features and original features and then convert them into a sequence.
We then use the mixture of experts to focus on the substructure of nodes under different relation combinations on the fraud graph, and a manager is developed to guide the training of experts and coordinate the detection results of different experts.
Finally, a regularization method for MoE is proposed to improve the robustness of the model.

In summary, our contributions are as follows:
\begin{itemize}
    \item {We propose a novel fraud detection framework that alleviates the negative effects of feature camouflage and relation camouflage simultaneously.}
    \item {We leverage mixture of experts to mitigate relation camouflage and propose a regularization method for MoE to improve the collaboration ability between experts.
    To our best knowledge, this is the first time MoE has been used for fraud detection task.}
    \item {We conducted extensive experiments and ablation studies to verify the effectiveness of SCFCRC on two real-world datasets. The results show that our approach has made significant progress on all experimental benchmarks, establishing a new state-of-the-art.
    Our code will be available at Github.
    }
\end{itemize}

\section{Related Works}

 \paragraph{GNN-based Fraud Detection.} Recently, many graph-based fraud detectors have been proposed and achieved promising results.
 CARE-GNN\cite{dou2020enhancing} and RioGNN\cite{peng2021reinforced} incorporate a GNN-enhanced neighbor selection module to tackle the camouflage problem in fraud detection. 
 To address the class imbalance in fraud detection scenarios, PC-GNN\cite{liu2021pick} introduces a node sampler and a label-aware neighbor selector to reweight the unbalanced classes.  
 H2-FDetector\cite{shi2022h2} identifies homogeneous and heterogeneous connections while applying separate aggregation strategies for different connection types respectively. 
 GTAN\cite{xiang2023semi} proposes an attribute-driven gated temporal attention network along with risk propagation for learning feature representation of transaction nodes. 
 These GNN-based methods are constrained by the conventional message transmission mode and are unable to effectively tackle the intricate issue of fraudster camouflage. 
 GAGA\cite{wang2023label} utilizes group aggregation to produce serialized neighborhood information, aggregates multi-hop domain messages using Transformer and enhances the original feature space with learnable encodings. 
 This brings valuable inspiration to our work.

 \paragraph{Mixture of Experts.} The Mixture-of-Experts\cite{jacobs1991adaptive} is an ensemble learning method first proposed in the machine learning community.
 To enhance the model’s ability to handle complex visual and speech data, DMoE\cite{eigen2013learning} extends the MoE structure to the deep neural networks, proposing a deep MoE model composed of multi-layer routers and experts. 
 Afterward, the MoE layers based on different basic neural network architectures have been proposed~\cite{shazeer2017outrageously,lepikhin2020gshard,xue2022go,zhou2022table}, achieving great success in various tasks.
 Specifically, it first decomposes a task into sub-tasks and then trains an expert model on each sub-task, a gating model is applied to learn which expert is competent and combine the predictions.
 In this paper, we apply MoE to fraud detection tasks, solving the camouflage problem to improve the performance of the model.

\section{Problem Statement}



We formulate the graph-based fraud detection as a semi-supervised node-level binary classification task.
Given a graph $\mathcal{G}=\{ \mathcal{V},\mathcal{E}, \mathcal{X},\mathcal{Y} \}$, $\mathcal{V}=\{ v_1,v_2,\ldots,v_N\}$ is the set containing N nodes, $\mathcal{E}=\{A^1,A^2,\ldots,A^R\}(R=|\mathcal{E}|)$ is the set of edges with a relation $r \in \{1,2,\ldots,R\}$.
$A$ represents the adjacency matrix, where $A^r_{uv}$ means that nodes $u$ and $v$ are connected under relation $r$.
$\mathcal{X}=\{ x_1,x_2,\ldots,x_N\}$ is the set of node features, $x_i \in \mathbb{R}^d$ is the $i^{th}$ node feature, $d$ is the dimension of feature.
In graph-based fraud detection, we consider a semi-supervised situation where only a portion of nodes $\hat{\mathcal{V}}$ in $\mathcal{G}$ are labeled.
Each node $v \in \hat{\mathcal{V}}$ has a corresponding binary label $y_v \in \{0,1\}$, where 0 represents benign nodes and 1 represents fraud, while other nodes remain unknown.
The number of classes is defined as $C=2$.

\section{Method}
In this section, we introduce the architecture of the proposed approach first and then present the details of the feature camouflage filter and the relation camouflage refiner. Finally, we introduce the loss function and the training process.

\subsection{Overview}

\begin{figure*}[tp]
\centering
\includegraphics[width=1.0\textwidth]{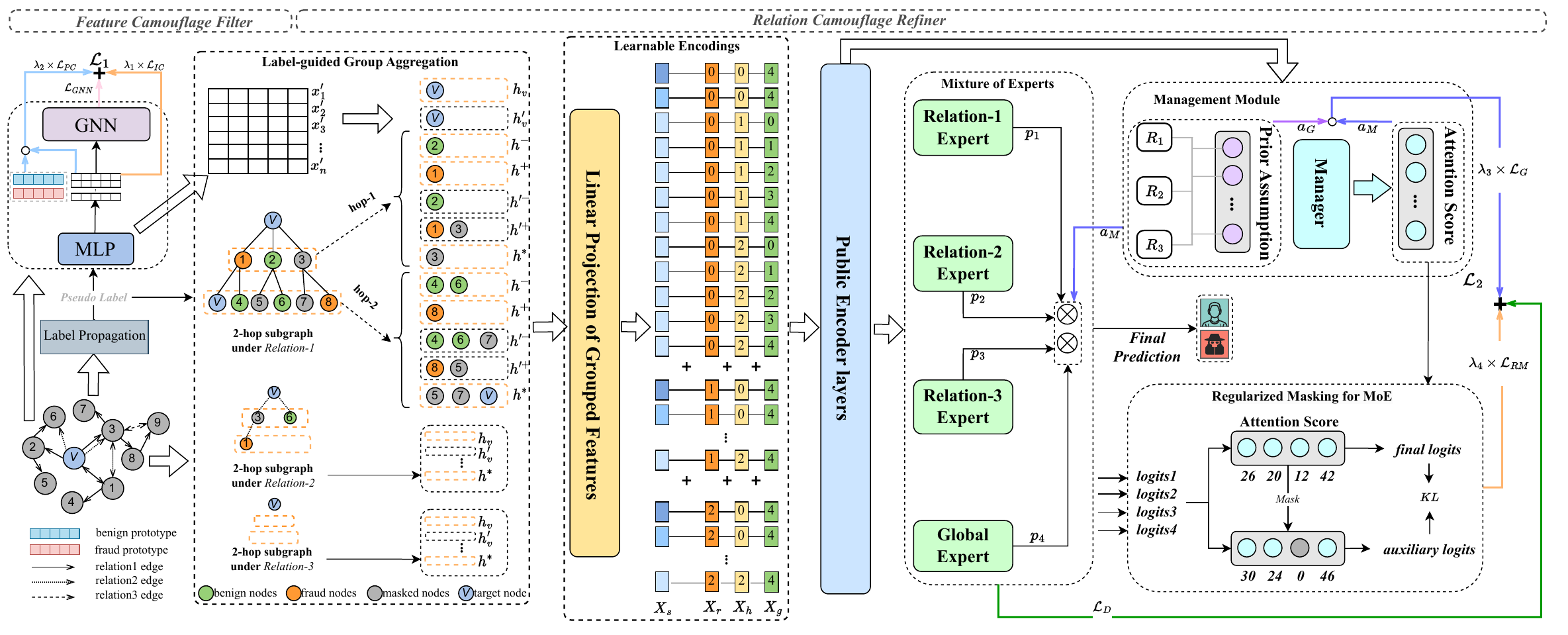}
\caption{
The overall architecture of SCFCRC.
For simplicity, the number of hops $K$ and relations $R$ are set to 2 and 3 respectively.
Four experts are used here, three experts focus on different relations and one global expert focuses on all relations.
}
\label{fig:framework}
\end{figure*}

The Figure~\ref{fig:framework} shows the architecture of SCFCRC. Generally, the model includes two components: Feature Camouflage Filter(FCF) and Relation Camouflage Refiner(RCR).
The feature camouflage filter focuses on the impact of structure on labels to avoid feature camouflage and outputs filtered features.
Relation camouflage refiner aims to refine relation camouflage, which consists of five modules:
1) The label-guided group aggregation generates a sequence of group vectors as the input based on original and filtered features.
2) In the learnable encodings module, the sequence of group vectors is encoded with three types of learnable embeddings.
3) The mixture of experts module for dealing with different relations of input.
4) The management module for guiding the training of experts and combining their ability of detection effectively.
5) The regularized masking for MoE is used to improve the robustness of MoE architectures.
Finally, we weightedly combine the outputs of different experts and generate the final prediction of the target node.

\subsection{Feature Camouflage Filter}
\label{subsec:Feature Camouflage Filter}
Fraudsters often disguise their true identities by mimicking the features of benign users.
Here, we propose the Feature Camouflage Filter, designed to remove the camouflage of features before they are fed into the classifier, to enhance fraud detection.
Details are as follows:

Formally, given a graph $\mathcal{G}$ with some nodes labeled, we perform label propagation on it, which ignores features and only considers the graph's structure.
Then we send the obtained pseudo label $L$ and graph $\mathcal{G}$ into the filter for training.
The filter consists of a Multi-Layer Perceptron(MLP) and a GNN, where the MLP is used to output filtered features $x'$:
\begin{equation}
    x' = MLP(x)
\end{equation}
The GNN is used to aggregate neighbor node information for graph representation learning.
The neighbor aggregation of node $v$ is as follows:
\begin{equation}
    \resizebox{.90\linewidth}{!}{$
            \displaystyle
            h_v^{(l)} = \sigma(W_g(h_v^{(l-1)} \oplus AGG^{(l)}\{h_u^{(l-1)},u \in \mathcal{N}^{(l)}(v)\}))
        $}
\end{equation}%
where $h_v^{(l-1)}$ and $h_u^{(l-1)}$ are the embedding of $v$ and $u$ at layer $l-1$, $h_v^{(0)}=x_v'$, a mean aggregator is used for all $AGG^{(l)}$, $W_g \in \mathbb{R}^{d_l \times d_{l-1}}$ is the parameter matrix, $\sigma$ is activation function, $\oplus$ denotes the embedding summation operation.
$\mathcal{N}^{(l)}(v)$ is the neighbor set of node $v$ at layer $l$

For each node $v$, its final embedding $z_v = h_v^{(L)}$ is the output of the GNN at the last layer $L$.
We use cross-entropy loss as the loss of GNN:
\begin{equation}
    \mathcal{L}_{GNN} = \sum_{v \in \mathcal{V}}^{} -\log (y_v\cdot \sigma(MLP(z_v))
\end{equation}

\subsubsection{Multi Contrastive Learning}
Inspired by the success of self-supervised contrastive learning~\cite{chen2020simple,he2020momentum}, we propose Multi Contrastive Learning to automatically learn how to align features of benign and fraudulent classes, which includes instance-wise and prototype-wise forms.
\paragraph{Instance-wise Contrastive Learning.} To make node features more discriminative, we use an instance-wise contrastive learning objective to cluster the same class and separate different classes of nodes, the objective function is computed as:
\begin{equation}
    \mathcal{L}_{IC} = -\frac{1}{|\mathcal{B}|}\sum_{i=1}^{|\mathcal{B}|}\sum_{\substack{j=1 \\ y_i=y_j}}^{|\mathcal{B}|}\log \frac{exp(cs(x'_i,x'_j))/\tau}{\sum_{k=1}^{|\mathcal{B}|}exp(cs(x'_i,x'_k)/\tau}
\end{equation}
where $|\mathcal{B}|$ is the number of nodes in a batch, $cs(\cdot)$ denotes the cosine similarity function and $\tau$ controls the temperature.

\paragraph{Prototype-wise Contrastive Learning.} In order to align the filtered feature space with the original feature and avoid excessive discrimination caused by instance-wise contrastive learning, we perform prototype-wise contrastive learning.
Prototypes $\{cen_j\}_{j=1}^C$ denote original feature averages with the same label in the batch data.
\begin{equation}
    \mathcal{L}_{PC} = -\frac{1}{|\mathcal{B}|}\sum_{i=1}^{|\mathcal{B}|}\sum_{\substack{j=1 \\ y_i=y_j}}^{C}\log \frac{exp(cs(x'_i,cen_j))/\tau}{\sum_{k=1}^{C}exp(cs(x'_i,cen_k)/\tau}
\end{equation}

\subsection{Relation Camouflage Refiner}
\label{subsec:Relation Camouflage Refiner}

\subsubsection{Label-guided Group Aggregation}
\label{subsec:Label-guided Group Aggregation}
Traditional graph neural network which relay on message passing mechanisms cannot effectively handle relation camouflage, so we devise a label-guided group(LGA) aggregation strategy to adapt filtered features.
Specifically, LGA treats filtered features as additional information and adds them to the aggregation of each hop based on the previous pseudo label.
Nodes with the same class label come into the same group and then each group performs aggregation separately:
\begin{equation}
   H_g^{\{k\}} = \lbrack h^-,h^+,h'^-,h'^+,h^* \rbrack^{\{k\}}
\end{equation}
\begin{equation}
     H_r=\mathop{\|}\limits_{k=1}^{K} H_g^{\{k\}}
\end{equation}
where $H_g$ is the sequence of group vectors, $h$ is the average aggregation of features after partitioning according to labels, 
$h^-$ for negative label nodes, $h^+$ for positive label nodes, $h'^-$ for pseudo-negative label nodes, $h^+$ for pseudo-positive label nodes,
and $h^*$ is the aggregation result of masked nodes whose labels are unknown.
$H_r$ is the group aggregation results of neighborhood information within $K$ hops under relation $r$, and $\mathop{\|}$ denotes concatenation operation.
We combine the raw feature $h_v$ and filtered feature $h'_v$ and the group vectors $H_r$ together into single sequence $H_{v,r} = \lbrack h_v \rbrack \mathop{\|} \lbrack h'_v \rbrack \mathop{\|} H_r$.
The next step is to combine all the sequences $H_{v,r}$ as the input feature sequence, which is defined as $H_{s} = \Vert_{r=1}^{R} H_{v,r}$.
Then we get the amount of vectors $S=R \times ((2\times C+1) \times K+2)$, where $S$ is the sequence length.

\subsubsection{Learnable Encodings}
To harness the structural, relational and label information in the original multi-relation fraud graphs, following previous work\cite{wang2023label}, we introduce three learnable encodings $X_r$, $X_h$, and $X_g$. The difference lies in the change is the length of the embedding sequence caused by the integration of the filtered feature.
\begin{equation}
   X_{in} = MLP(X_{s}) + X_{r} + X_{h} + X_{g}
\end{equation}
\begin{equation}
   \textbf{H} = f_{PEnc}(X_{in})
\end{equation}
where $X_{in}$ is the final sequence input, $\textbf{H} \in \mathbb{R}^{S \times d_{H}}$ denotes the representation vectors, $d_{H}$ represent the dimension of the encoder, $f_{PEnc}$ refers to public transformer encoder.

\subsubsection{Mixture of Experts Module}
A group of experts is applied to fraud detection, each of which is adept at processing different information of graph structure.
Experts share the same structure, which can be smoothly generalized to other scenarios, only the input dimensions accepted by the classifier may differ.
Specifically, each expert is implemented with a stack of transformer encoding layers and an MLP classifier that calculates the probability of fraud.
The process above is formulated as follows:
\begin{equation}
   \textbf{h}_i = AGG(f_{Enc_i}(\textbf{H}))
\end{equation}
\begin{equation}
   p_i = softmax(MLP_i(\textbf{h}_i))
\end{equation}
where $f_{Enc_i}$ is the $i^{th}$ expert's encoder, $\textbf{h}_i \in \mathbb{R}^{2 \times d_{H}}$ refers the aggregation of structural information that the $i^{th}$ expert is adept at processing.
$AGG$ is concat aggregator.
$p_i$ is the probability of fraud predicted by the $i^{th}$ expert, $MLP_i$ is the $i^{th}$ expert's classifier.

\subsubsection{Management Module}
This module consists of two components: structure perceptron and manager.
The structure perceptron is proposed to generate prior assumptions to guide the manager.
The manager is designed to guide experts' training and ensemble the results from all experts.

\paragraph{Structure Perceptron.} Fraudsters try to link multiple benign entities to disguise themselves.
According to previous work\cite{zheng2017smoke,kaghazgaran2018combating}, relation camouflage is usually established under a part of the relations of $\mathcal{E}$, not all of them.
Based on this, we propose the structure perceptron, which aims to generate a prior assumption under different structures, thereby guiding the manager to focus more on the structure in the non-camouflaged state.
Specifically, the structure perceptron module generates the prior assumption $a_G$ based on structures under different relations:
\begin{align}
   score_{r_i}(v) = \beta \cdot \frac{1}{|\mathcal{N}_{r_i}(v)|}\sum_{u \in \mathcal{N}_{r_i}(v)}^{} cs(x_u,x_v) + \nonumber \\
   (1-\beta) \cdot \frac{1}{|\mathcal{N}'_{r_i}(v)|}\sum_{u \in \mathcal{N}'_{r_i}(v)}^{} cs(x'_u,x'_v)
\end{align}%
where $score_{r_i}(v)$ and $\mathcal{N}_{r_i}(v)$ are respectively the score and neighbor set of node $v$ under relation set $r_i$ that the $i^{th}$ expert focuses on, $\beta$ is a hyperparameter.
We normalize $(score_{r_1}(v),score_{r_2}(v),\ldots,score_{r_n}(v))$ to get the final prior assumption $a_G$.

\paragraph{Manager.} We present a manager to guide the training of experts, which has the same network structure as experts.
The manager encodes $\textbf{H}$ and generates attention scores $a_M$:
\begin{equation}
   \textbf{h}_M = AGG(f_{Enc_M}(\textbf{H}))
\end{equation}
\begin{equation}
   a_M = softmax(MLP_M(\textbf{h}_M))
\end{equation}
where $\textbf{h}_M \in \mathbb{R}^{2 \times d_{H}}$ is representation of manager, $Enc_M$ and $MLP_M$ are the manager’s encoder and classifier respectively.

The prior assumption $a_G$ and attention score $a_M$ are used to teach the manager to allocate score reasonably and guide the expert's training,
the loss function $\mathcal{L}_G$, which calculates the logarithmic difference between the prior assumption $a_G$ and the attention scores $a_M$:
\begin{equation}
   \mathcal{L}_G = D_{KL}(a_G \Vert a_M)
\end{equation}
where $D_{KL}(\cdot\Vert\cdot)$ stands for the Kullback–Leibler divergence.

\subsubsection{Regularized Masking for MoE}
Previous MoE-based frameworks heavily relied on the manager's guidance during training and prediction.
However, the manager often failed to generate reasonable scores.
To address this, we proposed Regularized Masking for MoE(RMMoE).
RMMoE mitigates the impact of unreasonable allocation by randomly setting some expert scores to 0 during training and evenly redistributing these scores to unmasked experts, maintaining a total score of 1.
We use KL-divergence to constrain the output before and after the mask to be consistent:
\begin{equation}
   \mathcal{L}_{RM} = D_{KL}(\sum_{i=1}^{n_e}(a_M)_i*o_i \Vert \sum_{i=1}^{n_e}(a_{Mask})_i*o_i)
\end{equation}
where $n_e$ is the number of experts, $a_{Mask}$ is the attention score after performing the mask operation, $o_i$ is the output of the $i^{th}$ expert.

\subsection{Learning Objective}
\label{method:training}
Our training objective includes two parts.
First, We minimize the loss of feature camouflage filter $\mathcal{L}_1$:
\begin{equation}
   \mathcal{L}_{1} = \mathcal{L}_{GNN} +  \lambda_1\mathcal{L}_{ic} + \lambda_2\mathcal{L}_{pc}
\end{equation}
where $\lambda_1$ and $\lambda_2$ are hyperparameters that control the ratio of $L_{IC}$ and $L_{PC}$.

The second objective is to minimize the loss of relation camouflage refiner:
\begin{equation}
   \mathcal{L}_{D} = \sum_{i=1}^{n_e}(a_M)_i \cdot H_{CE}(p_i,y)
\end{equation}
\begin{equation}
   \mathcal{L}_{2} = \mathcal{L}_{D} +  \lambda_3\mathcal{L}_{G} + \lambda_4\mathcal{L}_{RM}
\end{equation}
where $L_D$ is the detection loss, $H_{CE}$ refers to the cross-entropy loss function.
$\lambda_3$ and $\lambda_4$ are hyperparameters that control the ratio of $L_{G}$ and $L_{RM}$.
We perform RMMoE in the training stationary phase and control it by using a hyperparameter $\delta$. When the training phase is less than $\delta$, the above loss becomes $\mathcal{L}_{2} = \mathcal{L}_{D} +  \lambda_3\mathcal{L}_{G}$.

\section{Experimentation}

\subsection{Experimental Setup}

\paragraph{Dataset.} We conduct experiments on two real-world fraud detection datasets to evaluate the effectiveness of our SCFCRC.
The statistic of datasets is shown in Table~\ref{tab:Statistics}.
\begin{itemize}
    \item \textbf{YelpChi}\cite{mcauley2013amateurs} includes hotel and restaurant reviews filtered (spam) and recommended (legitimate) by Yelp.
    The nodes in the YelpChi dataset are reviews with 32 handcrafted features, and the dataset includes three relations: R-U-R, R-S-R, and R-T-R.
    \item \textbf{Amazon}\cite{rayana2015collective} collects the product reviews of the Musical Instrument category on Amazon.com, in which nodes are users with 25 handcrafted features and the dataset encompasses three relations: U-P-U, U-S-U, and U-V-U.
\end{itemize}

\paragraph{Metrics.} 
We evaluate the detection performance with three widely used and complementary metrics: AUC, AP, and F1-macro. The AUC is the area under the ROC Curve that can evaluate the performance of classification by eliminating the influence of imbalanced classes. The AP is The Area Under the Precision Recall Curve, which focuses more on ranking fraudulent entities than benign ones. The F1-macro is the macroaverage of the two classes of F1 scores.
These metrics are bounded within the range [0, 1], with a higher value indicating superior performance. The results presented include the ten-run average and standard deviation obtained on the testing set.


\begin{table}[t]
\renewcommand{\arraystretch}{1.05}
\setlength{\tabcolsep}{1.2mm}
\centering
\begin{tabular}{c|ccccc}
\toprule
Dataset                  & \begin{tabular}[c]{@{}c@{}}\#Nodes \end{tabular}                   & IR                         & Relation & \#Relations   & \#Feat              \\ \midrule
\multirow{3}{*}{YelpChi} & \multirow{3}{*}{\begin{tabular}[c]{@{}c@{}}45,954\end{tabular}}    & \multirow{3}{*}{5.9}        & R-U-R    & 49,315        & \multirow{3}{*}{32}  \\
                         &                                                                    &                            & R-S-R    & 3,402,743     &                     \\
                         &                                                                    &                            & R-T-R    & 573,616       &                     \\ \bottomrule
\multirow{3}{*}{Amazon}  & \multirow{3}{*}{\begin{tabular}[c]{@{}c@{}}11,944\end{tabular}}    & \multirow{3}{*}{13.5}      & U-P-U    & 175,608       & \multirow{3}{*}{25} \\
                         &                                                                    &                            & U-S-U    & 3,566,479     &                     \\
                         &                                                                    &                            & U-V-U    & 1,036,737     &                     \\ \midrule
\end{tabular}
\caption{The Statistic of Datasets. IR represents the class imbalance ratio.}
\label{tab:Statistics}
\end{table}

\paragraph{Baselines.} We chose some traditional and improved GNNs as the baseline: GCN\cite{kipf2016semi}, GAT\cite{velickovic2017graph}, HAN\cite{wang2019heterogeneous}, GraphSAGE\cite{hamilton2017inductive}, GraphSAINT\cite{zeng2019graphsaint}, Cluster-GCN\cite{chiang2019cluster}, and SIGN\cite{frasca2020sign}. Besides, some state-of-the-art methods for graph-based fraud detection were used to compare with our approach as follows: CARE-GNN\cite{dou2020enhancing}, RioGNN\cite{peng2021reinforced}, PC-GNN\cite{liu2021pick}, FRAUDRE\cite{zhang2021fraudre}, H2-FDetector\cite{shi2022h2}, GTAN\cite{xiang2023semi} and GAGA\cite{wang2023label}. 
In the classical GNN model, the multi-relation graph is amalgamated into a homogeneous graph. We select parameters based on the content of relevant papers.

\begin{table*}[ht]
\renewcommand{\arraystretch}{1.05}
\centering
\begin{tabular*}{1.0\linewidth}{l|lll|lll}
\toprule
\multirow{2}{*}{Methods} & \multicolumn{3}{c|}{YelpChi}                                                                                & \multicolumn{3}{c}{Amazon}                                                                                \\ \cline{2-7}
                         & \multicolumn{1}{c}{AUC}           & \multicolumn{1}{c}{AP}            & \multicolumn{1}{c|}{F1-macro}       & \multicolumn{1}{c}{AUC}           & \multicolumn{1}{c}{AP}            & \multicolumn{1}{c}{F1-macro}      \\ \midrule
GAN                      & \multicolumn{1}{c}{0.5924±0.0030} & \multicolumn{1}{c}{0.2176±0.0119} & \multicolumn{1}{c|}{0.5072±0.0271}  & \multicolumn{1}{c}{0.8405±0.0075} & \multicolumn{1}{c}{0.4660±0.0131} & \multicolumn{1}{c}{0.6985±0.0046} \\
GAT                      & \multicolumn{1}{c}{0.6796±0.0070} & \multicolumn{1}{c}{0.2807±0.0048} & \multicolumn{1}{c|}{0.5773±0.0080} & \multicolumn{1}{c}{0.8096±0.0113} & \multicolumn{1}{c}{0.3082±0.0067} & \multicolumn{1}{c}{0.6681±0.0076} \\
HAN                      & \multicolumn{1}{c}{0.7420±0.0009} & \multicolumn{1}{c}{0.2722±0.0036} & \multicolumn{1}{c|}{0.5472±0.0097}  & \multicolumn{1}{c}{0.8421±0.0062} & \multicolumn{1}{c}{0.4631±0.0185} & \multicolumn{1}{c}{0.7016±0.0126} \\ \midrule
GraphSAGE                & 0.7409±0.0000                     & \multicolumn{1}{r}{0.3258±0.0000} & \multicolumn{1}{r|}{0.6001±0.0002}  & \multicolumn{1}{r}{0.9172±0.0001} & 0.8268±0.0002                     & 0.9029±0.0004                     \\
Cluster-GCN              & 0.7623±0.0069                     & \multicolumn{1}{r}{0.3691±0.0179} & \multicolumn{1}{r|}{0.6204±0.0557}  & \multicolumn{1}{r}{0.9211±0.0256} & 0.8075±0.0566                     & 0.8853±0.0272                     \\
GraphSAINT               & 0.7412±0.0143                     & \multicolumn{1}{r}{0.3641±0.0304} & \multicolumn{1}{r|}{0.5974±0.0728}  & \multicolumn{1}{r}{0.8946±0.0176} & 0.7956±0.0091                     & 0.8888±0.0244                     \\ \midrule
CARE-GNN                 & 0.7854±0.0111                     & 0.3972±0.0208                     & 0.6064±0.0186                       & 0.8823±0.0305                     & 0.7609±0.0904                     & 0.8592±0.0574                     \\
FRAUDRE                  & 0.7588±0.0078                     & 0.3870±0.0186                     & 0.6421±0.0135                       & 0.9308±0.0180                     & 0.8433±0.0089                     & 0.9037±0.0031                     \\
PC-GNN                   & 0.8154±0.0031                     & 0.4797±0.0064                     & 0.6523±0.0197                       & 0.9489±0.0067                     & 0.8435±0.0166                     & 0.8897±0.0144                     \\
RioGNN                   & 0.8144±0.0050                     & 0.4722±0.0079                     & 0.6422±0.0233                       & 0.9558±0.0019                     & 0.8700±0.0044                     & 0.8848±0.0125                     \\
H2-FDetector             & 0.8892±0.0020                     & 0.5543±0.0135                     & 0.7345±0.0086                       & 0.9605±0.0008                     & 0.8494±0.0023                     & 0.8010±0.0058                     \\
GTAN                     & 0.8987±0.0023                     & 0.6891±0.0089                     & 0.7816±0.0077                       & 0.9441±0.0316                     & 0.8561±0.0429                     & 0.9105±0.0096                     \\ \midrule
SIGN                     & 0.8569±0.0051                     & 0.5801±0.0191                     & 0.7308±0.0053                       & 0.9404±0.0033                     & 0.8483±0.0031                     & 0.9046±0.0012                     \\ \midrule
GAGA                     & 0.9439±0.0016                     & 0.8014±0.0063                     & 0.8323±0.0041                       & 0.9629±0.0052                     & 0.8815±0.0095                     & 0.9133±0.0040                     \\ \midrule
- w/o FCF               & 0.9485±0.0028                     & 0.8183±0.0086                     & 0.8462±0.0047                        & 0.9594±0.0055                     & 0.8781±0.0044                     & 0.9261±0.0010                     \\
- w/o RCR               & 0.9493±0.0027                     & 0.8215±0.0083                     & 0.8481±0.0064                        & 0.9611±0.0028                     & 0.8759±0.0087                     & 0.9241±0.0022                      \\ \midrule
Ours                     & \textbf{0.9539±0.0033}                     & \textbf{0.8348±0.0110}                     & \textbf{0.8560±0.0069}                       & \textbf{0.9627±0.0010}                     & \textbf{0.8901±0.0036}                     &            \textbf{0.9279±0.0029}                       \\ \bottomrule
\end{tabular*}
\caption{Performance Comparison on YelpChi and Amazon.}
\label{tab:overall Performance}
\end{table*}

\begin{table}[ht]
\centering
\renewcommand{\arraystretch}{1.05}
\begin{tabular}{l|cc}
\toprule
\multirow{2}{*}{Model} & \multicolumn{2}{c}{F1-macro}                       \\ \cline{2-3} 
                       & \multicolumn{1}{c|}{YelpChi}       & Amazon        \\ \midrule
Ours                   & \multicolumn{1}{c|}{\textbf{0.8560±0.0069}} & \textbf{0.9279±0.0029}  \\ \midrule
- w/o $\mathcal{L}_{IC}$                   & \multicolumn{1}{c|}{0.8527±0.0035} & 0.9254±0.0014 \\ \midrule
- w/o $\mathcal{L}_{PC}$                   & \multicolumn{1}{c|}{0.8550±0.0014} & 0.9270±0.0037 \\ \midrule
- w/o $\mathcal{L}_G$                  & \multicolumn{1}{c|}{0.8521±0.0064} & 0.9233±0.0014 \\ \midrule
- w/ fixed $a_G$                   & \multicolumn{1}{c|}{0.8544±0.0036} & 0.9255±0.0024 \\ \midrule
- w/o $\mathcal{L}_{RM}$                  & \multicolumn{1}{c|}{0.8510±0.0076} & 0.9260±0.0020 \\ \bottomrule
\end{tabular}
\caption{
Ablation study on YelpChi and Amazon.
It shows the results of training without $\mathcal{L}_{IC}$, $\mathcal{L}_{PC}$, $\mathcal{L}_G$, $\mathcal{L}_{RM}$ and with a fixed prior assumption $a_G$.
}
\label{tab:ablation study}
\end{table}


\paragraph{Experiment Settings.} Our implementations are based on Pytorch and DGL.
The size of the training/validation/testing set for all compared methods is set to 0.4/0.1/0.5.
The hyperparameters $\lambda_1$/$\lambda_2$/$\lambda_3$/$\lambda_4$ that control the loss weight are set to 0.1/0.1/0.1/0.3.
$\beta$ is set to 0.5.
The masking ratio on YelpChi and Amazon are set to 0.15 and 0.1 respectively.
$\delta$ is set to 0.4.
In this paper, we set the number of experts to 4 and the hop count is set to 2.
The embedding size on YelpChi and Amazon are set to 32 and 16 respectively, batch size is 512 and 256.
The number of layers for each expert and gate is set to 1.
The number of public layers on YelpChi and Amazon is set to 2 and 1 respectively.
To avoid overfitting, we use dropout mechanism with a dropout rate 0.1.
The learning rate is set to 3e-3 and weight decay is 1e-4.
All methods are optimized with Adam optimizer.

\subsection{Overall Detection Results}
The overall performance is shown in Table~\ref{tab:overall Performance}.
First of all, it is observed that almost all general GNN models perform weaker than enhanced graph-based fraud detection models. The reason for this is that those general GNN models are based on the homogeneity assumption and cannot handle the noises introduced by camouflage. But SIGN has achieved some performance improvement by aggregating multi-hop neighborhood information, with AUC close to PC-GNN on Amazon and even better performance on YelpChi.
This highlights the beneficial nature of multi-hop neighborhood information.
Compared to state-of-the-art graph-based fraud detection methods, our proposal significantly outperforms them.
Among them, GTAN uses risk embedding and propagation while randomly masking risk features during training, and GAGA introduces a group aggregation module to generate distinguishable multi-hop neighborhood information. Compared with other graph-based fraud detection methods, these two methods make full use of the label information while learning multi-hop neighborhood information, and have substantial improvement overall. 
But compared with GAGA, SCFCRC achieves better performance improvement on both datasets and achieves 2.37\%, 3.34\%, and 1\% gains in F1-macro, AP, and AUC on YelpChi. The reason behind this can be attributed to the fact that in addition to learning multi-hop neighborhood information and label information, SCFCRC also solves the feature and relation camouflage problem. Overall, the performance shows the effectiveness of SCFCRC.

\begin{figure*}[ht]
\centering  
\subfigure[Original]{
\label{Fig.sub.0}
\includegraphics[width=0.19\textwidth]{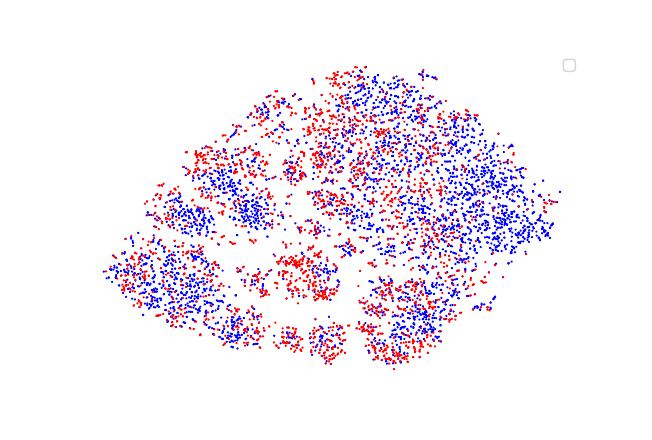}}
\subfigure[Without contrast]{
\label{Fig.sub.1}
\includegraphics[width=0.19\textwidth]{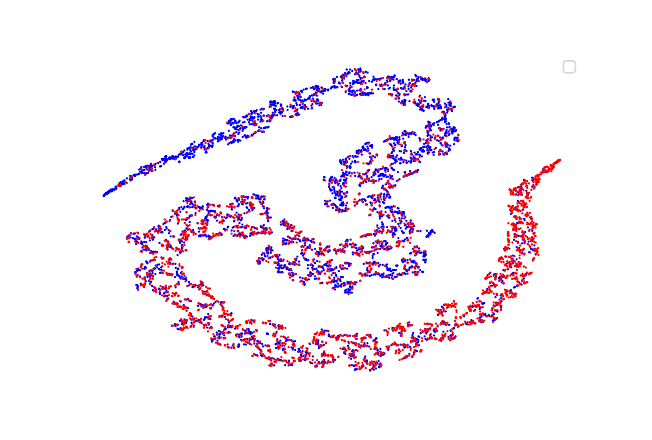}}
\subfigure[$\mathcal{L}_{IC}$]{
\label{Fig.sub.2}
\includegraphics[width=0.19\textwidth]{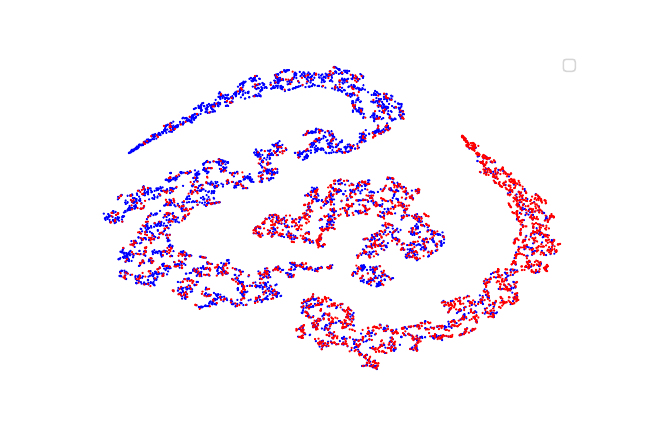}}
\subfigure[$\mathcal{L}_{PC}$]{
\label{Fig.sub.3}
\includegraphics[width=0.19\textwidth]{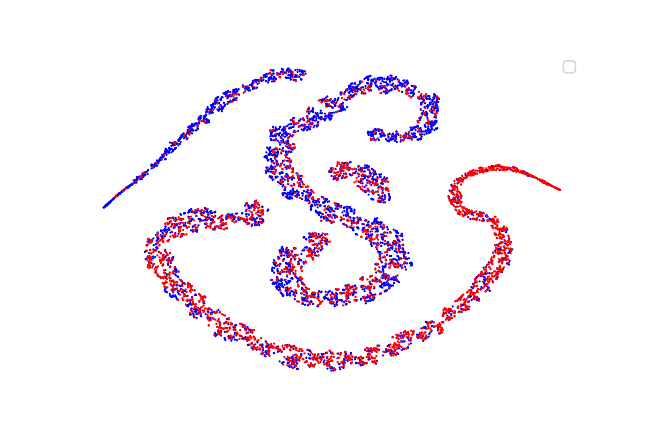}}
\subfigure[$\mathcal{L}_{IC}+\mathcal{L}_{PC}$]{
\label{Fig.sub.4}
\includegraphics[width=0.19\textwidth]{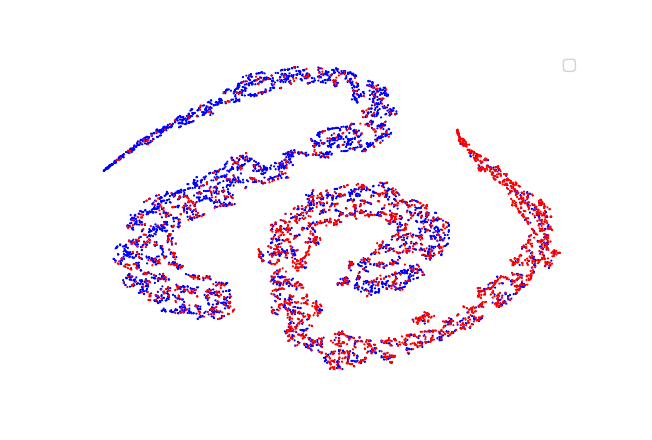}}
\caption{The t-SNE visualization of features on YelpChi. (Benign: blue, Fraud: red)}
\label{fig:Visualization}
\end{figure*}

\begin{figure}[ht]
\centering  
\subfigure[Trained with $\mathcal{L}_D$]{
\label{Fig.sub.x}
\includegraphics[width=0.22\textwidth]{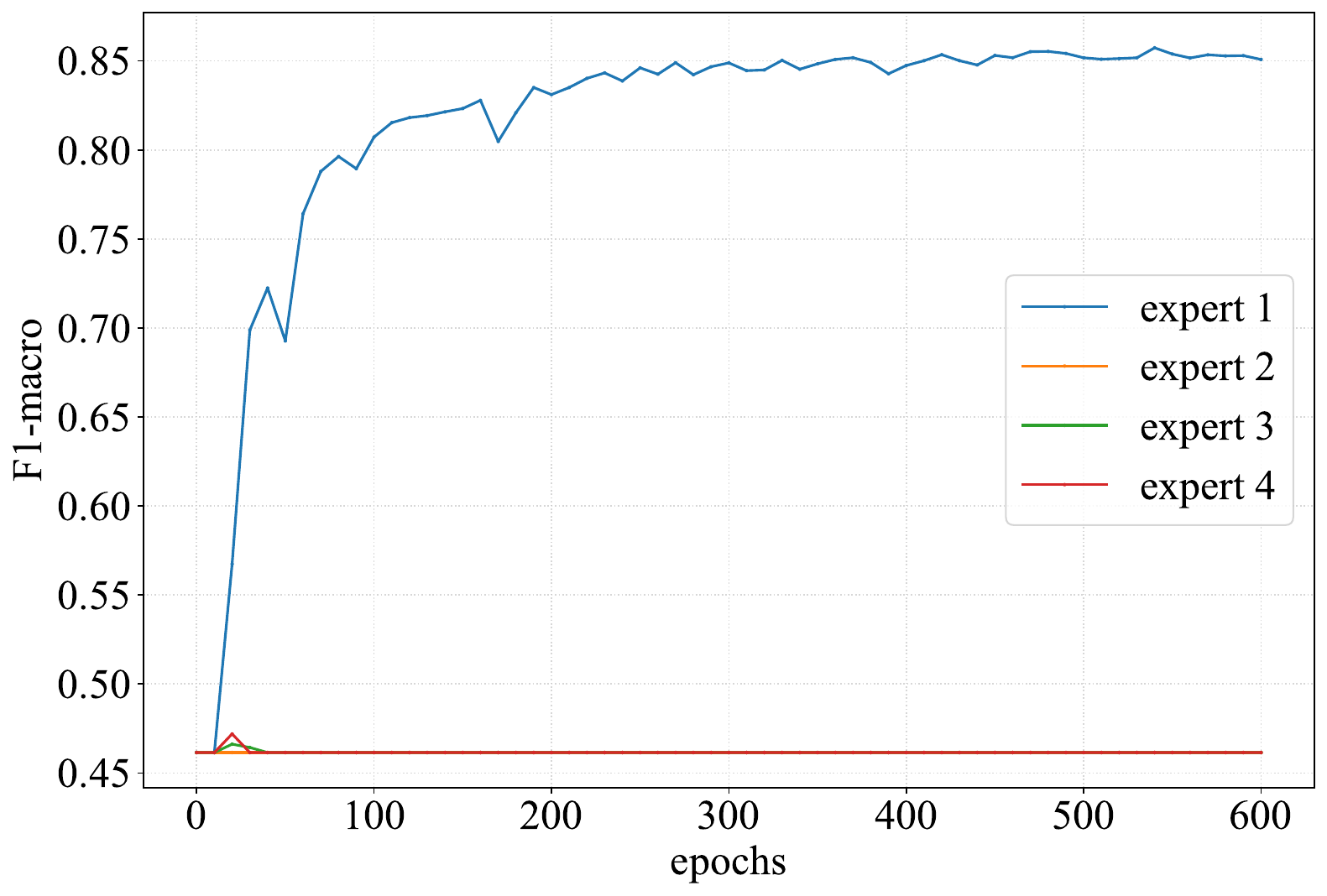}}
\subfigure[Trained with $\mathcal{L}_D+\mathcal{L}_G$]{
\label{Fig.sub.y}
\includegraphics[width=0.22\textwidth]{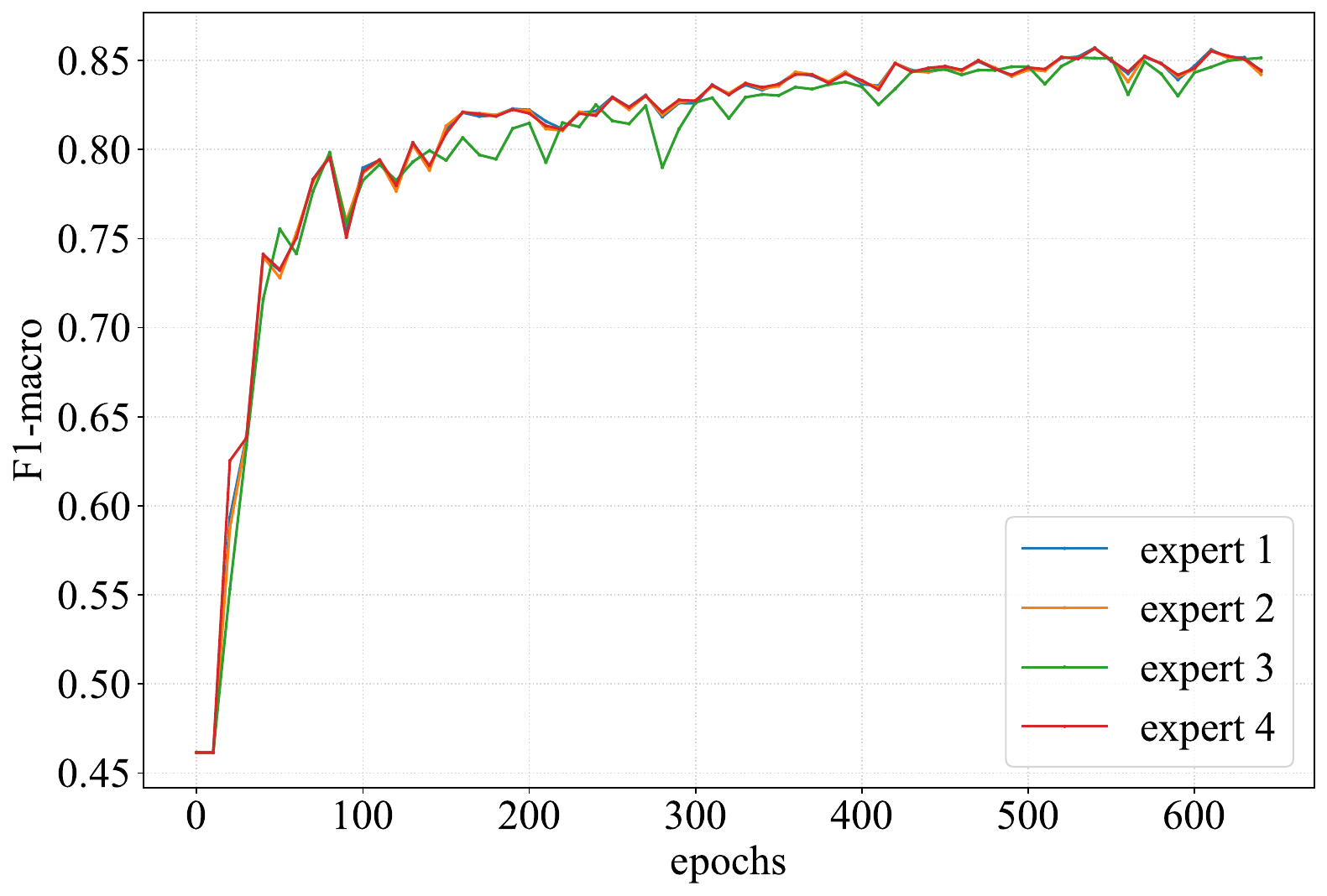}}
\caption{Comparison of models trained with/without the manager guidance loss on YelpChi.}
\label{fig:expert differentiation}
\end{figure}

\subsection{Ablation Study}
To assess the impact of counteracting the two camouflages on our model's performance, we conducted an ablation study by individually removing the FCF and RCR.
The results are shown in Table \ref{tab:overall Performance}.
We can observe that the performance of the two variants decreased significantly.
First, When FCF is removed, the performance drops by 0.54\%, 1.65\%, and 0.98\% in terms of AUC, AP, and F1-macro on YelpChi.
This demonstrates that FCF can mitigate feature camouflage and supplement more available information.
Second, When RCR is removed, we can observe that the performance drops by 0.46\% in AUC, 1.33\% in AP, and 0.79\% in F1-macro on YelpChi.
This shows that RCR can effectively refine complex relations structures into multiple relatively simple structures, thereby alleviating the problem of relation camouflage.
In addition, Table \ref{tab:ablation study} shows more detailed ablation experiments.

\textit{w/o $\mathcal{L}_{IC}$}: 
When instance-wise contrastive learning is removed, the distinction between features of different classes decreases, resulting in a decrease in detection performance.

\textit{w/o $\mathcal{L}_{PC}$}:
The ablation experimental results verify the effectiveness of prototype-wise contrastive learning.
Detailed visualization will be discussed in next section.

\textit{w/o $\mathcal{L}_G$}:
We conduct an ablation study on both datasets without the guidance loss $\mathcal{L}_G$, leading to a significant drop in performance.
The "imbalanced experts" phenomenon will be discussed in more detail in a subsequent section.

\textit{w/ fixed $a_G$}:
We initialize the previous hypothesis $a_G$ as $(0.2,0.2,0.2,0.4)$ and do not use the score generated by the structure perceptron.
F1-macro on YelpChi and Amazon has dropped by 0.16\% and 0.24\% respectively.

\textit{w/o $\mathcal{L}_{RM}$}:
The ablation experimental results verify the effectiveness of regularized masking for MoE, and the mask ratio experiment will be discussed in subsequent section.

\subsection{Visualization}
\label{subsec:visualization}
We visualize node features after FCF using various training methods, using the YelpChi dataset as an example.
We employ the t-SNE\cite{van2008visualizing} to map the vector representation into the 2-dimensional space.
Due to the extreme class imbalance, we undersample the benign class to keep the number of both classes close for convenient visualization.
As shown in Figure \ref{fig:Visualization}, we can observe that the original node representation distribution is messy.
Without contrastive learning, the training method shows observable clustering in node representations, but fails to distinguish fraud nodes from benign nodes.
Instance-wise contrastive learning can significantly improve this phenomenon.
Compared with IC, prototype-wise contrastive learning makes the intra-category distance more compact, but may cause multiple clusters.
Finally, we use both contrastive learning methods simultaneously, producing stronger intra-class cohesion and inter-class segregation, generating notably better embeddings.

\begin{figure}[t]
\centering  
\subfigure[YelpChi]{
\label{Fig.sub.masking.a}
\includegraphics[width=0.22\textwidth]{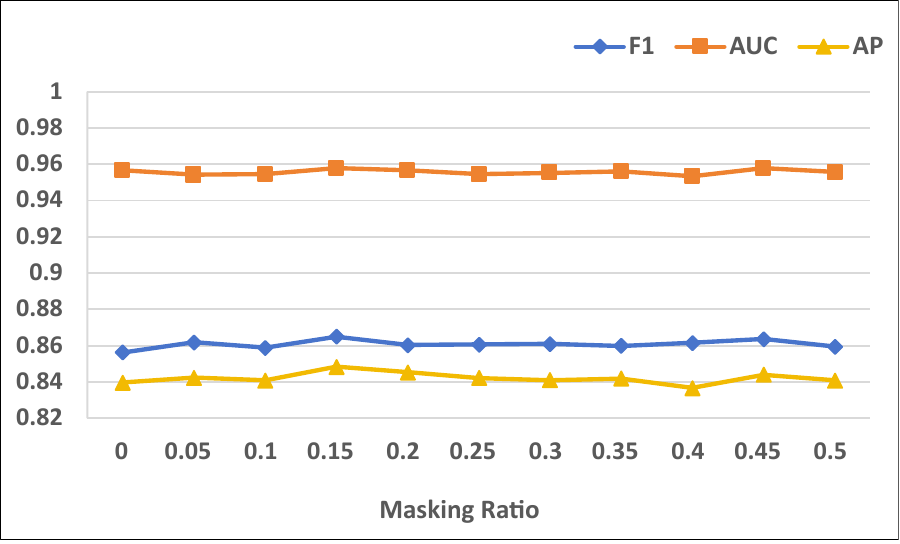}}
\subfigure[Amazon]{
\label{Fig.sub.masking.b}
\includegraphics[width=0.22\textwidth]{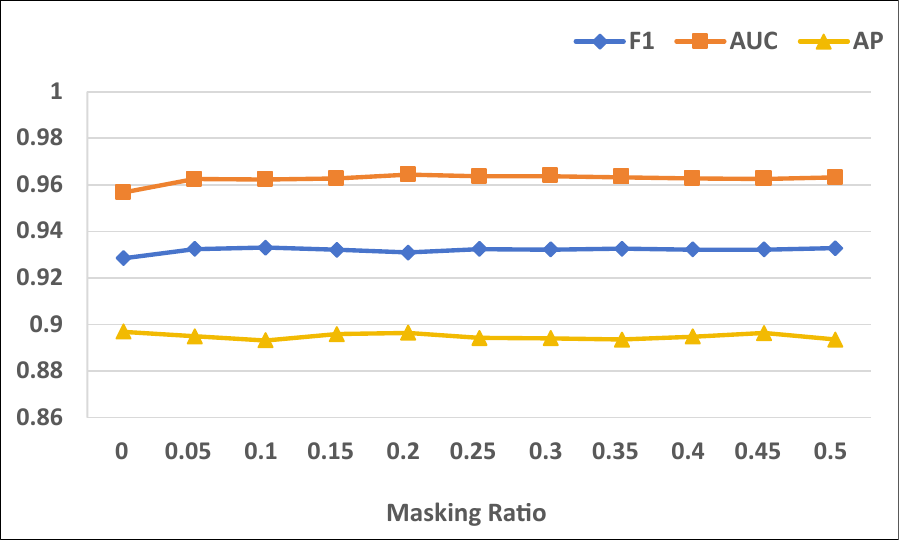}}
\caption{Performance with different masking ratio on both datasets.}
\label{fig:RM}
\end{figure}

\subsection{Analyzing Experts Differentiation}
\label{subsec:imbalanced experts}
To validate the effectiveness of the manager, we investigated the differentiation of experts, which refers to the model’s ability to achieve balanced training across experts based on the proposed manager assumption loss $\mathcal{L}_G$.
Figure \ref{fig:expert differentiation} presents a comparison of two models that were trained with and without $\mathcal{L}_G$, depicting the performance curves of different experts in each subgraph on the YelpChi dataset.
As shown in Figure \ref{Fig.sub.y}, each expert is well-trained, and the phenomenon of "imbalanced experts" does not occur.
However, Figure \ref{Fig.sub.x} shows that when trained without employing $\mathcal{L}_G$, only one expert is effectively trained, while the performance of the other three experts remains stagnant at around 0.46 despite increasing training steps. This is far less effective than the complete model.
The results show that the manager achieves balanced training among experts and significantly raises the performance upper bound.

\subsection{Effects of Regularized Masking for MoE}
\label{subsec:masking ratio}
The regularized masking for MoE can enhance the model’s robustness and collaboration among experts while reducing the impact of improper allocation of scores by the manager.
The masking ratio is an important parameter. Figure \ref{fig:RM} illustrates the change in masking ratio from 0 to 0.5 and the corresponding variations in F1, AP, and AUC values on two datasets.
It is observed that the improvement is no longer significant when the masking ratio exceeds 0.1 and 0.15 in the two datasets respectively.
But it is still better than the result without using RMMoE(masking ratio is 0).
This proves the validity of our proposed regularized masking method.

\section{Conclusion}
In this paper, we proposed a novel approach to detect fraudsters in the multi-relational graph setting with counteracting camouflage behaviors. In particular, we use label propagation to generate pseudo-labels, which are combined with contrastive learning to obtain camouflage-removed features. To address the relation camouflage, we use Mixture-of-Experts and regularized masking to enhance model robustness.
Comprehensive experiments show our proposed method significantly outperforms the state-of-the-art on two public fraud detection datasets.
In future work, we plan to explore removing both types of camouflage in an iterative manner, which we believe could achieve more refined decamouflage effects.


\end{document}